\documentclass[10pt,twocolumn]{article}

\usepackage[letterpaper,top=0.78in,bottom=0.82in,left=0.72in,right=0.72in,columnsep=0.24in]{geometry}
\usepackage{microtype}
\usepackage{graphicx}
\usepackage{booktabs}
\usepackage{amsmath}
\usepackage{amssymb}
\usepackage{mathtools}
\usepackage{algorithm}
\usepackage{algorithmic}
\usepackage{array}
\usepackage{multirow}
\usepackage{makecell}
\usepackage{xcolor}
\usepackage[numbers,sort&compress]{natbib}
\usepackage[hidelinks]{hyperref}
\usepackage[capitalize,noabbrev]{cleveref}

\title{\textbf{ChainDoRA: Tensor-Train Factorized Weight-Decomposed Low-Rank
Adaptation for Parameter-Efficient LLM Fine-Tuning}}

\author{
Ashfak Yeafi$^{1}$ \and
Mehedi Hasan$^{2}$ \and
Md Khairul Islam$^{3}$\\[1mm]
\small $^{1}$Department of Electrical and Electronic Engineering, Khulna University of Engineering \& Technology, Khulna, Bangladesh\\
\small $^{2}$Department of Computer Science and Engineering, Brac University, Dhaka, Bangladesh\\
\small $^{3}$Department of Mathematics and Computer Science, Hobart and William Smith Colleges, Geneva, NY, USA\\[1mm]
\small \texttt{yeafiashfak@gmail.com} \quad
\texttt{mehedi.hasan1@g.bracu.ac.bd} \quad
\texttt{khairul.robotics@gmail.com}
}

\date{}

\begin{document}
\maketitle

\begin{abstract}
Parameter-efficient fine-tuning (PEFT) adapts large language models (LLMs)
with a small task-specific parameter budget, but the dense low-rank factors
used by LoRA and retained in DoRA still scale linearly with the dimensions of
each adapted layer. We propose \emph{ChainDoRA}, which preserves DoRA's
magnitude--direction decomposition while constructing its directional
low-rank factors from a connected Tensor-Train (TT) chain whose boundary rank
is the adapter rank and whose independent internal TT rank controls capacity
and parameter cost. Under a controlled 15,119-example response-only
adaptation setting with LLaMA-7B, ChainDoRA with TT rank 16 achieves 72.30\%
average accuracy across seven commonsense reasoning benchmarks, compared with
69.88\% for LoRA and 69.39\% for DoRA. At the same adapter rank, ChainDoRA
uses 5.35M trainable parameters versus 56.10M for LoRA and 56.98M for DoRA,
a 90.62\% reduction relative to DoRA. Ablations over TT rank and adapter
placement reveal controllable parameter--accuracy trade-offs, while
cross-architecture analysis shows that the structural parameter savings
persist across representative 3B--72B model configurations. Code is available
at \url{https://github.com/AshfakYeafi/Chain-DoRA}.
\end{abstract}

\noindent\textbf{Keywords:} parameter-efficient fine-tuning; low-rank adaptation;
DoRA; Tensor-Train decomposition; large language models; model adaptation.

\section{Introduction}
\label{sec:introduction}

The rapid growth of pretrained language models has made task-specific adaptation increasingly expensive in terms of memory, optimization, and model storage. Conventional full-parameter fine-tuning updates the entire pretrained model and consequently requires maintaining a complete task-specific parameter set. Parameter-efficient fine-tuning (PEFT) addresses this limitation by keeping most or all pretrained parameters frozen while optimizing a comparatively small set of task-specific parameters. Representative approaches include lightweight adapter modules~\cite{houlsby2019parameter}, continuous prefix optimization~\cite{li2021prefix}, and low-rank reparameterization~\cite{hu2022lora}. Among these methods, Low-Rank Adaptation (LoRA) has become particularly attractive because its learned update can be merged into the pretrained weight at inference time without introducing an additional inference branch.

LoRA parameterizes the update to a frozen weight matrix through two trainable low-rank factors~\cite{hu2022lora}. This substantially reduces the number of optimized parameters relative to full fine-tuning, but its adapter size remains proportional to the input and output dimensions of every adapted linear transformation. Moreover, LoRA modifies the pretrained weight additively without explicitly separating changes in weight scale from changes in orientation. Weight-Decomposed Low-Rank Adaptation (DoRA) addresses the latter issue by decomposing the pretrained weight into magnitude and directional components and applying LoRA to the directional component while learning the magnitude separately~\cite{liu2024dora}. DoRA has demonstrated improved adaptation capability over LoRA across language and vision-language tasks; however, its directional update continues to rely on the same pair of densely parameterized low-rank matrices used by LoRA.

Tensor decompositions offer an alternative mechanism for representing high-dimensional parameters through structured low-dimensional factors. The Tensor-Train (TT) format represents a high-order tensor as a sequence of low-order cores connected through auxiliary TT ranks~\cite{oseledets2011tensor}. TT representations have previously been used to compress neural-network parameters~\cite{novikov2015tensorizing}, and recent work has extended tensorized parameterizations to PEFT. LoRETTA employs TT representations for ultra-low-parameter LLM adaptation~\cite{yang2024loretta}, while TT-LoRA explores TT-based parameterization for efficient large-model adaptation~\cite{anjum2024ttlora}. More recent approaches have also investigated weight-decomposed tensor adaptation and unified TT representations for constructing correlated low-rank factors. These studies indicate that tensor-network structure can substantially reduce the number of directly optimized parameters, but they also highlight the importance of how the tensor representation is integrated into the adaptation mechanism.

Motivated by these observations, we propose \emph{ChainDoRA}, a Tensor-Train-factorized extension of magnitude--direction adaptation. Rather than directly optimizing the two dense low-rank matrices in the directional branch of DoRA, ChainDoRA represents them through contractions of a single connected sequence of trainable TT cores. The input-side cores contract to form
$A_{\mathrm{TT}}\in\mathbb{R}^{r\times d_{\mathrm{in}}}$, while the output-side cores form
$B_{\mathrm{TT}}\in\mathbb{R}^{d_{\mathrm{out}}\times r}$.
The two contractions are connected through a TT bond whose dimension is fixed to the adapter rank $r$. A separate internal TT rank $\rho$ controls the capacity of the tensorized representation. Consequently, the nominal low-rank adaptation dimension and the internal tensor complexity can be controlled independently.

\begin{figure*}[t]
    \centering
    \includegraphics[width=0.88\textwidth]{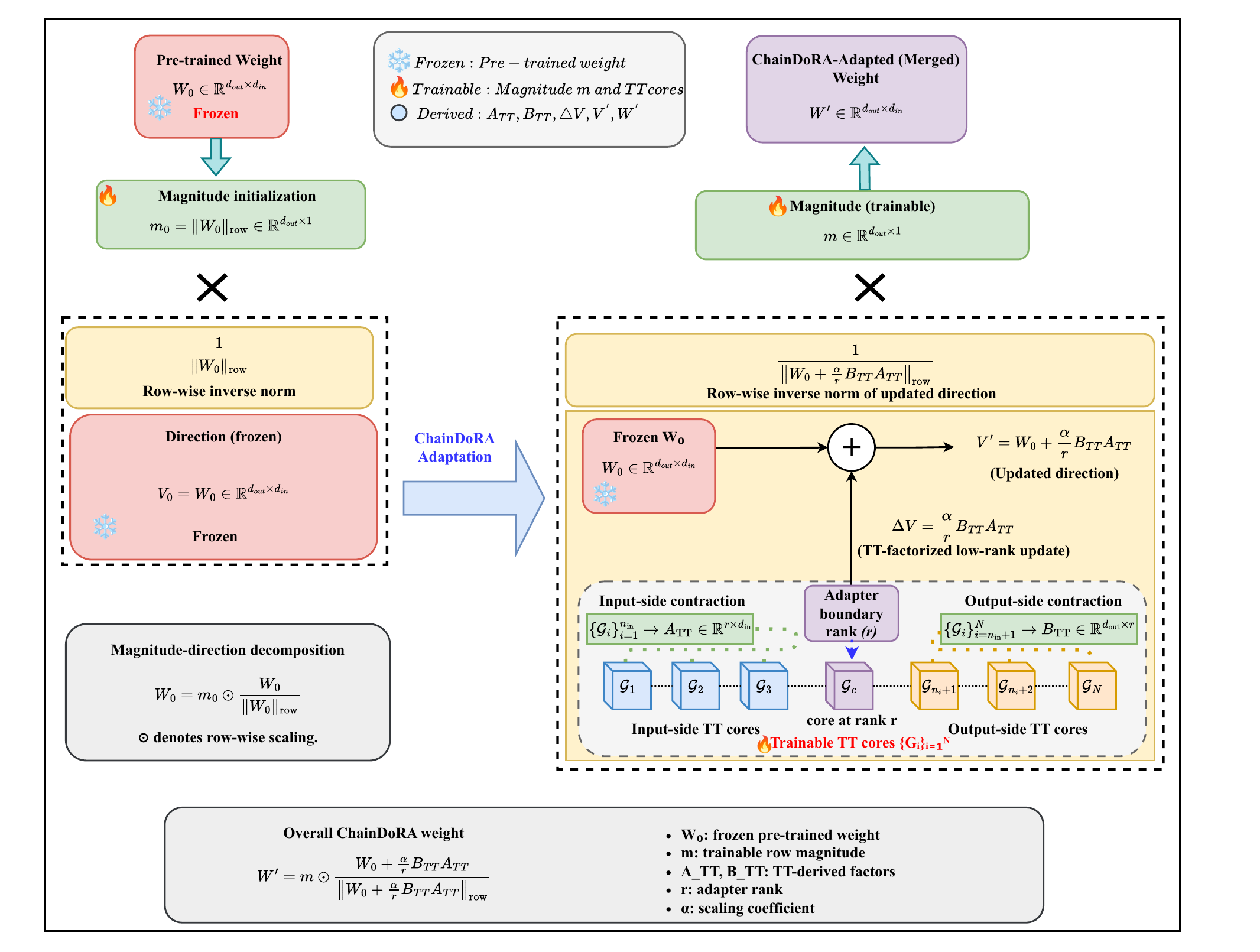}
    \caption{Overview of the proposed ChainDoRA framework. A connected
    Tensor-Train chain constructs the directional factors
    $A_{\mathrm{TT}}$ and $B_{\mathrm{TT}}$, which are coupled through the
    adapter-rank boundary $r$. Their product defines the low-rank directional
    update, while the row-wise magnitude vector $m$ is optimized independently.
    The pretrained weight $W_0$ remains frozen.}
    \label{fig:chaindora_overview}
\end{figure*}

As illustrated in Fig.~\ref{fig:chaindora_overview}, the resulting TT-derived update is incorporated into the magnitude--direction decomposition rather than used as an independent additive tensor adapter. This design preserves the row-wise magnitude control of DoRA while replacing its dense directional factors with a structured tensor representation. For the LLaMA-7B configuration studied in this work, adapting the query, key, value, up-projection, and down-projection layers with rank $r=32$ requires approximately 56.98M trainable parameters with DoRA. ChainDoRA with internal TT rank $\rho=16$ reduces this number to 5.35M, corresponding to a 90.62\% reduction in trainable parameters while retaining the same nominal adapter rank.

We evaluate ChainDoRA under a controlled 15,119-example response-only adaptation setting with LLaMA-7B as the frozen backbone. The evaluation covers seven commonsense reasoning benchmarks: BoolQ, PIQA, SocialIQA, WinoGrande, ARC-Easy, ARC-Challenge, and OpenBookQA. Under matched optimization, target-module, and evaluation settings, ChainDoRA achieves a seven-task average accuracy of 72.30\%, compared with 69.88\% for LoRA and 69.39\% for DoRA. We further investigate the effects of TT rank and adapter placement, demonstrating a tunable parameter--accuracy trade-off. In addition, we analyze ChainDoRA's parameter scaling across multiple LLM architectures and model sizes to characterize how the proposed factorization scales beyond LLaMA-7B.

The main contributions of this work are summarized as follows:
\begin{itemize}

    \item We introduce \emph{ChainDoRA}, a magnitude--direction PEFT framework that replaces the densely parameterized directional factors of DoRA with factors obtained from a connected Tensor-Train representation.

    \item We formulate the adapter rank $r$ as the boundary rank connecting the input- and output-side TT contractions while introducing an independent internal TT rank $\rho$. This separates the nominal low-rank adaptation dimension from the parameter capacity of the tensor representation.

    \item We develop a parameter-efficient implementation in which the TT-derived directional update is integrated with row-wise magnitude normalization, together with a balanced nonzero TT initialization designed for stable optimization of the multiplicative tensor chain.

    \item We conduct a controlled comparison with LoRA and DoRA on seven commonsense reasoning benchmarks using LLaMA-7B. ChainDoRA achieves a 72.30\% average accuracy with 5.35M trainable parameters, improving over both baselines while reducing the trainable parameter count by approximately an order of magnitude.

    \item We provide systematic ablations of TT rank and adaptation location, together with a cross-architecture parameter-scaling analysis, to characterize the efficiency--capacity trade-offs of the proposed parameterization.

\end{itemize}

The remainder of this paper is organized as follows. Section~\ref{sec:related_work} reviews low-rank, weight-decomposed, and tensor-based PEFT methods. Section~\ref{sec:methodology} presents the ChainDoRA formulation, TT construction, initialization strategy, and parameter complexity. Section~\ref{sec:experiments} describes the experimental protocol. The empirical results and ablation studies are then presented and discussed, followed by the conclusion.

\section{Related Work}
\label{sec:related_work}

\subsection{Parameter-Efficient and Low-Rank Adaptation}
\label{subsec:related_peft}

Parameter-efficient fine-tuning (PEFT) adapts pretrained models while updating
only a small subset of parameters, thereby reducing the memory and storage
requirements associated with conventional full-parameter fine-tuning.
Early PEFT approaches introduced lightweight trainable modules between frozen
Transformer layers~\cite{houlsby2019parameter}, while prefix tuning optimized
continuous task-specific vectors that are prepended to the hidden states of
Transformer layers~\cite{li2021prefix}. These approaches established that
effective downstream adaptation does not necessarily require updating the
entire pretrained parameter set.

Low-Rank Adaptation (LoRA)~\cite{hu2022lora} instead reparameterizes the
update of a frozen weight matrix $W_0$ as the product of two trainable
low-rank matrices. For a rank-$r$ adapter, the update takes the form
$BA$, where
$A\in\mathbb{R}^{r\times d_{\mathrm{in}}}$ and
$B\in\mathbb{R}^{d_{\mathrm{out}}\times r}$.
This formulation substantially reduces the number of trainable parameters
relative to full fine-tuning and allows the learned update to be merged into
the pretrained weight for inference. LoRA has subsequently served as the
basis for numerous PEFT variants that modify rank allocation, parameter
sharing, initialization, or backbone precision.

AdaLoRA~\cite{zhang2023adalora}, for example, parameterizes incremental
updates using a singular-value-decomposition-inspired formulation and
dynamically reallocates the available rank budget according to parameter
importance. QLoRA~\cite{dettmers2023qlora} addresses a complementary source
of memory cost by backpropagating through a frozen 4-bit quantized backbone
while optimizing LoRA adapters. VeRA~\cite{kopiczko2024vera} further reduces
task-specific storage by sharing a pair of frozen random low-rank matrices
across layers and learning comparatively small scaling vectors. These methods
demonstrate that parameter efficiency can be improved through rank
allocation, quantization, and parameter sharing; nevertheless, conventional
LoRA and many of its derivatives retain an explicitly matrix-factorized
representation of the task-specific update.

\subsection{Magnitude--Direction Adaptation}
\label{subsec:related_dora}

Weight-Decomposed Low-Rank Adaptation (DoRA)~\cite{liu2024dora} analyzes
fine-tuning through the decomposition of a pretrained weight into magnitude
and directional components. Instead of applying a low-rank update directly
to the complete weight matrix, DoRA learns the magnitude independently and
uses LoRA to adapt the directional component. The resulting weight can be
written conceptually as a learned magnitude multiplied by a normalized,
low-rank-updated direction. This decomposition was motivated by observed
differences between the magnitude and directional changes produced by full
fine-tuning and conventional LoRA. DoRA was shown to improve over LoRA across
language, vision-language, and multimodal adaptation tasks while preserving
the ability to merge the adapted weights for inference~\cite{liu2024dora}.

Despite this decomposition, the directional branch of DoRA is still
parameterized through the two dense rank-$r$ factors inherited from LoRA.
For an adapted matrix with dimensions
$d_{\mathrm{out}}\times d_{\mathrm{in}}$, these factors require
$r(d_{\mathrm{in}}+d_{\mathrm{out}})$ trainable parameters before accounting
for the magnitude vector. Consequently, the directional adapter size
continues to grow linearly with the dimensions of the target transformation.
ChainDoRA retains the magnitude--direction decomposition of DoRA but replaces
these directly optimized directional matrices with factors derived from a
structured Tensor-Train representation.

\subsection{Tensorized Parameter-Efficient Fine-Tuning}
\label{subsec:related_tensor_peft}

Tensor decomposition provides a structured alternative to conventional matrix
factorization for representing high-dimensional parameters. The
Tensor-Train (TT) format introduced by Oseledets~\cite{oseledets2011tensor}
represents a high-order tensor through a sequence of low-order cores connected
by auxiliary ranks. The number of parameters is determined by the physical
mode sizes and TT ranks rather than by explicitly storing the complete tensor.
TT representations were subsequently applied to neural networks by
tensorizing dense weight matrices and replacing them with compact TT
structures~\cite{novikov2015tensorizing}.

Recent work has extended tensor-network representations to PEFT.
LoRETTA~\cite{yang2024loretta} proposes two TT-based adaptation strategies.
LoRETTA$_{\mathrm{adp}}$ tensorizes lightweight adapter modules, whereas
LoRETTA$_{\mathrm{rep}}$ performs weight reparameterization using compact
TT factors. The latter demonstrates that Tensor-Train structure can provide
substantial parameter reductions during LLM fine-tuning. TT-LoRA
~\cite{anjum2024ttlora} further investigates Tensor-Train parameterization
for LLM adaptation and studies the trade-off among tensorization structure,
TT rank, parameter count, and downstream performance. In contrast to
standard LoRA, these approaches use higher-order tensor structure to reduce
the number of explicitly optimized adaptation parameters.

Other tensorized methods adopt different decompositions or initialization
strategies. DoTA~\cite{hu2024dota} introduces Weight-Decomposed Tensor
Adaptation based on a Matrix Product Operator (MPO) representation and
leverages a decomposition of pretrained weights to obtain an informative
initialization for tensor adaptation. Its quantized extension, QDoTA, combines
this formulation with low-bit backbone quantization. Although DoTA and
ChainDoRA both involve weight-decomposed adaptation, their parameterizations
are fundamentally different: DoTA adapts weights through an MPO-based tensor
representation initialized from pretrained weights, whereas ChainDoRA retains
the DoRA magnitude--direction formulation and tensorizes the low-rank
directional factors.

TensorGuide~\cite{qi2025tensorguide} is particularly related to the present
work because it considers the interaction between the two low-rank matrices
rather than treating them as independent objects. TensorGuide employs a
unified TT network driven by controlled Gaussian inputs to jointly generate
correlated low-rank adaptation matrices. This contrasts with separately
tensorizing the two LoRA factors and demonstrates that coupling their
parameterization can improve the efficiency and expressivity of tensorized
low-rank adaptation.

ChainDoRA shares the general motivation of exploiting dependencies between
low-rank factors but differs in both construction and placement of the tensor
representation. Specifically, ChainDoRA forms a single connected sequence of
trainable TT cores whose input-side and output-side contractions directly
produce
$A_{\mathrm{TT}}\in\mathbb{R}^{r\times d_{\mathrm{in}}}$ and
$B_{\mathrm{TT}}\in\mathbb{R}^{d_{\mathrm{out}}\times r}$.
The TT bond separating the two contractions is explicitly assigned the
adapter rank $r$, while the remaining bonds are governed by an independent
TT rank $\rho$. The resulting product
$B_{\mathrm{TT}}A_{\mathrm{TT}}$ is then used specifically to modify the
directional component of a DoRA-style magnitude--direction decomposition.
Thus, the adapter rank controlling the effective low-rank update and the TT
rank controlling the internal tensor capacity are explicitly separated in
the proposed formulation.

\begin{table*}[t]
\centering
\caption{Conceptual comparison of representative low-rank and tensorized
parameter-efficient adaptation methods. ``Joint $A/B$ structure'' indicates
whether the two low-rank factors are generated or parameterized through a
shared structured representation.}
\label{tab:related_method_comparison}
\resizebox{\textwidth}{!}{
\begin{tabular}{lccccc}
\toprule
\textbf{Method}
&
\textbf{Low-Rank Update}
&
\textbf{Tensor Representation}
&
\textbf{Magnitude--Direction}
&
\textbf{Joint $A/B$ Structure}
&
\textbf{Primary Structural Mechanism}
\\
\midrule

LoRA~\cite{hu2022lora}
&
\checkmark
&
--
&
--
&
--
&
Dense trainable low-rank factors
\\

AdaLoRA~\cite{zhang2023adalora}
&
\checkmark
&
--
&
--
&
--
&
Adaptive rank-budget allocation
\\

DoRA~\cite{liu2024dora}
&
\checkmark
&
--
&
\checkmark
&
--
&
Magnitude--direction decomposition
\\

LoRETTA~\cite{yang2024loretta}
&
\checkmark
&
TT
&
--
&
--
&
Tensorized adapters / reparameterization
\\

TT-LoRA~\cite{anjum2024ttlora}
&
\checkmark
&
TT
&
--
&
--
&
TT-parameterized adaptation
\\

DoTA~\cite{hu2024dota}
&
--
&
MPO
&
Weight-decomposed
&
--
&
Pretrained-weight MPO initialization
\\

TensorGuide~\cite{qi2025tensorguide}
&
\checkmark
&
TT
&
--
&
\checkmark
&
Unified TT-generated correlated factors
\\

\textbf{ChainDoRA}
&
\checkmark
&
\textbf{TT}
&
\checkmark
&
\checkmark
&
\textbf{Connected TT directional factors with boundary rank $r$}
\\

\bottomrule
\end{tabular}}
\end{table*}

Table~\ref{tab:related_method_comparison} summarizes these distinctions.
Rather than treating tensor decomposition as an independent adapter or as a
replacement for the complete adapted weight, ChainDoRA uses a connected TT
representation specifically within the directional branch of
magnitude--direction adaptation. This construction motivates the formulation
developed in Section~\ref{sec:methodology}.

\section{Methodology}
\label{sec:methodology}

The proposed ChainDoRA framework combines the magnitude--direction
reparameterization of DoRA~\cite{liu2024dora} with a connected
Tensor-Train (TT) representation~\cite{oseledets2011tensor} of the
directional low-rank factors. Instead of directly optimizing the two dense
low-rank matrices used by LoRA and DoRA, ChainDoRA constructs these factors
from contractions of a single TT chain. The adapter rank $r$ is assigned to
the TT bond separating the input-side and output-side contractions, while a
distinct internal TT rank $\rho$ controls the capacity and parameter cost of
the tensorized representation. The pretrained weight remains frozen, and only
the TT cores and the row-wise magnitude parameters are optimized.

\begin{figure}[t]
    \centering
    \includegraphics[
        width=\columnwidth
    ]{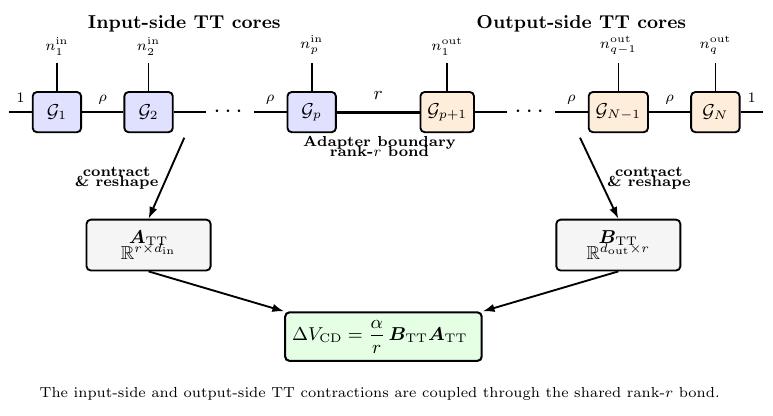}
    \caption{Connected Tensor-Train construction of ChainDoRA. The
    input-side and output-side TT contractions generate
    $A_{\mathrm{TT}}$ and $B_{\mathrm{TT}}$, respectively. The adapter
    rank $r$ forms the boundary bond between the two contractions, while
    the remaining internal TT bonds are controlled by rank $\rho$.}
    \label{fig:tt_chain}
\end{figure}

Fig.~\ref{fig:chaindora_overview} provides the overall architecture, while
Fig.~\ref{fig:tt_chain} illustrates the internal TT construction used to
generate the two directional factors.


\subsection{Preliminaries: LoRA and Weight-Decomposed Adaptation}
\label{subsec:preliminaries}

Consider a pretrained linear transformation with frozen weight

\begin{equation}
W_0
\in
\mathbb{R}^{d_{\mathrm{out}}\times d_{\mathrm{in}}},
\label{eq:base_weight}
\end{equation}

where $d_{\mathrm{in}}$ and $d_{\mathrm{out}}$ denote the input and output
dimensions, respectively. LoRA~\cite{hu2022lora} represents a task-specific
weight update as

\begin{equation}
\Delta W_{\mathrm{LoRA}}
=
sBA,
\qquad
s=\frac{\alpha}{r},
\label{eq:lora_update_method}
\end{equation}

where

\begin{equation}
A\in\mathbb{R}^{r\times d_{\mathrm{in}}},
\qquad
B\in\mathbb{R}^{d_{\mathrm{out}}\times r}.
\label{eq:lora_factor_dimensions}
\end{equation}

Here, $r$ is the low-rank adaptation dimension and $\alpha$ is the LoRA
scaling coefficient. Since

\begin{equation}
\operatorname{rank}(BA)\leq r,
\label{eq:lora_rank_bound}
\end{equation}

the update matrix has rank at most $r$, reducing the number of trainable
degrees of freedom relative to a dense update while the pretrained matrix
$W_0$ remains unchanged.

DoRA~\cite{liu2024dora} further decomposes adaptation into magnitude and
direction. For the
$d_{\mathrm{out}}\times d_{\mathrm{in}}$ weight-storage convention used in
our implementation, the decomposition is performed row-wise. The initial
magnitude is therefore defined as

\begin{equation}
m_i^{(0)}
=
\left\|W_{0,i:}\right\|_2,
\qquad
i=1,\ldots,d_{\mathrm{out}},
\label{eq:magnitude_initialization}
\end{equation}

or equivalently,

\begin{equation}
m^{(0)}
=
\left\|W_0\right\|_{\mathrm{row}}.
\label{eq:magnitude_vector}
\end{equation}

Given an unnormalized directional matrix $V$, the corresponding
magnitude--direction reconstruction is

\begin{equation}
W
=
m\odot
\frac{V}
{\left\|V\right\|_{\mathrm{row}}},
\label{eq:weight_decomposition}
\end{equation}

where $\odot$ denotes row-wise broadcasting. In conventional DoRA, the
directional matrix is updated using the LoRA factors,

\begin{equation}
V_{\mathrm{DoRA}}
=
W_0+sBA.
\label{eq:dora_direction}
\end{equation}

ChainDoRA preserves the decomposition in
(\ref{eq:weight_decomposition}) but replaces the directly parameterized
matrices $A$ and $B$ with factors generated from a connected Tensor-Train
representation.


\subsection{Connected Tensor-Train Directional Factorization}
\label{subsec:connected_tt}

Let the input and output dimensions be factorized into smaller physical modes,

\begin{equation}
d_{\mathrm{in}}
=
\prod_{j=1}^{p} n_j^{\mathrm{in}},
\qquad
d_{\mathrm{out}}
=
\prod_{j=1}^{q} n_j^{\mathrm{out}},
\label{eq:tensorized_dimensions}
\end{equation}

where $p$ and $q$ denote the numbers of input-side and output-side modes,
respectively. These modes are concatenated into a single sequence

\begin{equation}
(n_1,\ldots,n_N)
=
(
n_1^{\mathrm{in}},\ldots,n_p^{\mathrm{in}},
n_1^{\mathrm{out}},\ldots,n_q^{\mathrm{out}}
),
\qquad
N=p+q.
\label{eq:combined_modes}
\end{equation}

Following the TT representation~\cite{oseledets2011tensor}, ChainDoRA
introduces $N$ trainable third-order cores

\begin{equation}
\mathcal{G}_j
\in
\mathbb{R}^{R_{j-1}\times n_j\times R_j},
\qquad
j=1,\ldots,N,
\label{eq:tt_core_definition}
\end{equation}

with boundary ranks

\begin{equation}
R_0=R_N=1.
\label{eq:tt_outer_ranks}
\end{equation}

A central design choice of ChainDoRA is to assign the TT bond between the
input-side and output-side contractions directly to the adapter rank,

\begin{equation}
R_p=r.
\label{eq:adapter_boundary_rank}
\end{equation}

The remaining non-boundary TT ranks are controlled by a separate internal
rank $\rho$,

\begin{equation}
R_j
=
\begin{cases}
1,
& j=0 \ \text{or}\ j=N,\\
r,
& j=p,\\
\rho,
& \text{otherwise}.
\end{cases}
\label{eq:chaindora_rank_structure}
\end{equation}

Consequently, the complete rank sequence can be written compactly as

\begin{equation}
\mathbf{R}
=
\left(
1,
\underbrace{\rho,\ldots,\rho}_{p-1},
r,
\underbrace{\rho,\ldots,\rho}_{q-1},
1
\right).
\label{eq:rank_vector}
\end{equation}

The role of $r$ is therefore distinct from that of $\rho$: $r$ controls the
maximum rank of the resulting directional update, whereas $\rho$ controls the
capacity of the TT representation used to construct its factors.

For a fixed physical index $i_j$, define the corresponding TT-core matrix
slice as

\begin{equation}
G_j^{(i_j)}
=
\mathcal{G}_j(:,i_j,:)
\in
\mathbb{R}^{R_{j-1}\times R_j}.
\label{eq:tt_slice}
\end{equation}

Let $\phi_{\mathrm{in}}(\cdot)$ and
$\phi_{\mathrm{out}}(\cdot)$ denote the flattening maps from the tensorized
input and output indices to the original matrix coordinates. The contraction
of the first $p$ cores yields a $1\times r$ vector for each input coordinate.
The corresponding column of $A_{\mathrm{TT}}$ is

\begin{equation}
A_{\mathrm{TT}}
\left[
:,
\phi_{\mathrm{in}}(i_1,\ldots,i_p)
\right]
=
\left(
G_1^{(i_1)}
G_2^{(i_2)}
\cdots
G_p^{(i_p)}
\right)^{\mathrm{T}}.
\label{eq:att_contraction}
\end{equation}

Similarly, contraction of the output-side cores yields an $r\times1$ vector,
and the corresponding row of $B_{\mathrm{TT}}$ is

\begin{equation}
B_{\mathrm{TT}}
\left[
\phi_{\mathrm{out}}(o_1,\ldots,o_q),
:
\right]
=
\left(
G_{p+1}^{(o_1)}
\cdots
G_N^{(o_q)}
\right)^{\mathrm{T}}.
\label{eq:btt_contraction}
\end{equation}

Therefore,

\begin{equation}
A_{\mathrm{TT}}
\in
\mathbb{R}^{r\times d_{\mathrm{in}}},
\qquad
B_{\mathrm{TT}}
\in
\mathbb{R}^{d_{\mathrm{out}}\times r}.
\label{eq:tt_factor_dimensions}
\end{equation}

Importantly, $A_{\mathrm{TT}}$ and $B_{\mathrm{TT}}$ are not stored as
independent trainable dense matrices. Their entries are differentiable
functions of the TT cores on the two sides of the shared rank-$r$ boundary.

The ChainDoRA directional residual is defined as

\begin{equation}
\Delta V_{\mathrm{CD}}
=
sB_{\mathrm{TT}}A_{\mathrm{TT}},
\qquad
s=\frac{\alpha}{r}.
\label{eq:chaindora_directional_update}
\end{equation}

Since the intermediate dimension is still $r$,

\begin{equation}
\operatorname{rank}
\left(
\Delta V_{\mathrm{CD}}
\right)
\leq r,
\label{eq:chaindora_rank_bound}
\end{equation}

regardless of the internal TT rank $\rho$. Increasing $\rho$ therefore
increases the representational capacity of the TT factors without changing
the nominal upper bound on the rank of the final directional update.


\subsection{Magnitude--Direction Composition}
\label{subsec:chaindora_composition}

The unnormalized ChainDoRA directional matrix is

\begin{equation}
V_{\mathrm{CD}}
=
W_0
+
sB_{\mathrm{TT}}A_{\mathrm{TT}}.
\label{eq:chaindora_v}
\end{equation}

The effective adapted weight is obtained through row-wise normalization and
magnitude scaling,

\begin{equation}
W_{\mathrm{CD}}
=
m
\odot
\frac{
V_{\mathrm{CD}}
}{
\left\|
V_{\mathrm{CD}}
\right\|_{\mathrm{row}}
}.
\label{eq:chaindora_weight}
\end{equation}

Substituting (\ref{eq:chaindora_v}) gives the complete ChainDoRA
reparameterization,

\begin{equation}
W_{\mathrm{CD}}
=
m\odot
\frac{
W_0+sB_{\mathrm{TT}}A_{\mathrm{TT}}
}{
\left\|
W_0+sB_{\mathrm{TT}}A_{\mathrm{TT}}
\right\|_{\mathrm{row}}
}.
\label{eq:chaindora_complete}
\end{equation}

The set of trainable parameters of an adapted layer is therefore

\begin{equation}
\Theta_{\mathrm{CD}}
=
\left\{
m,
\mathcal{G}_1,
\ldots,
\mathcal{G}_N
\right\},
\label{eq:trainable_set}
\end{equation}

while $W_0$ remains frozen throughout fine-tuning.


\subsection{Factored Row-Norm Computation}
\label{subsec:factored_norm}

Direct evaluation of
$\|W_0+sB_{\mathrm{TT}}A_{\mathrm{TT}}\|_{\mathrm{row}}$
by first constructing the dense
$d_{\mathrm{out}}\times d_{\mathrm{in}}$ matrix
$B_{\mathrm{TT}}A_{\mathrm{TT}}$ introduces unnecessary temporary storage.
The required norm can instead be evaluated through low-rank intermediate
quantities. A related factored-norm formulation has recently been studied for
scaling DoRA implementations~\cite{zelenin2026scalingdora}.

For notational simplicity, let

\begin{equation}
A=A_{\mathrm{TT}},
\qquad
B=B_{\mathrm{TT}}.
\label{eq:ab_short}
\end{equation}

For output row $i$,

\begin{equation}
\begin{aligned}
\left\|
W_{0,i:}+sB_{i:}A
\right\|_2^2
&=
\left\|W_{0,i:}\right\|_2^2
\\
&\quad+
2s
\left\langle
W_{0,i:},
B_{i:}A
\right\rangle
\\
&\quad+
s^2
\left\|
B_{i:}A
\right\|_2^2.
\end{aligned}
\label{eq:norm_expansion}
\end{equation}

Define

\begin{equation}
q_i
=
\left\|W_{0,i:}\right\|_2^2,
\label{eq:q_term}
\end{equation}

\begin{equation}
c_i
=
\sum_{k=1}^{r}
B_{ik}
\left(
W_0A^{\mathrm{T}}
\right)_{ik},
\label{eq:cross_term}
\end{equation}

and

\begin{equation}
u_i
=
B_{i:}
\left(
AA^{\mathrm{T}}
\right)
B_{i:}^{\mathrm{T}}.
\label{eq:gram_term}
\end{equation}

The row norm is then

\begin{equation}
n_i
=
\sqrt{
q_i
+
2sc_i
+
s^2u_i
+
\epsilon
},
\label{eq:factored_row_norm}
\end{equation}

where $\epsilon$ is a small numerical-stability constant. The dominant
rank-dependent intermediate matrices are

\begin{equation}
W_0A^{\mathrm{T}}
\in
\mathbb{R}^{d_{\mathrm{out}}\times r}
\label{eq:cross_intermediate}
\end{equation}

and

\begin{equation}
AA^{\mathrm{T}}
\in
\mathbb{R}^{r\times r},
\label{eq:gram_intermediate}
\end{equation}

rather than the full
$d_{\mathrm{out}}\times d_{\mathrm{in}}$ matrix $BA$.

The linear action of the low-rank residual can likewise be evaluated
sequentially. For an input $x$,

\begin{equation}
z
=
A_{\mathrm{TT}}x,
\qquad
z\in\mathbb{R}^{r},
\label{eq:low_rank_projection}
\end{equation}

followed by

\begin{equation}
\Delta y
=
B_{\mathrm{TT}}z.
\label{eq:low_rank_expansion}
\end{equation}

Let

\begin{equation}
g_i
=
\frac{m_i}{n_i}.
\label{eq:row_gain}
\end{equation}

The complete forward transformation can therefore be expressed as

\begin{equation}
y
=
g
\odot
\left[
W_0x
+
sB_{\mathrm{TT}}
\left(
A_{\mathrm{TT}}x
\right)
\right].
\label{eq:chaindora_forward}
\end{equation}

This formulation avoids materializing
$B_{\mathrm{TT}}A_{\mathrm{TT}}$ solely for either the linear residual or the
row-norm computation.


\subsection{Detached Direction Normalization}
\label{subsec:dora_simple}

The original DoRA formulation notes that retaining the normalization term in
the complete gradient graph introduces additional training-memory
overhead~\cite{liu2024dora}. Following its low-memory modification, our
implementation treats the dynamically computed directional norm as a constant
during backpropagation.

Let

\begin{equation}
\bar{n}
=
\operatorname{stopgrad}
\left(
\left\|
V_{\mathrm{CD}}
\right\|_{\mathrm{row}}
\right).
\label{eq:detached_norm}
\end{equation}

The training-time reparameterization is therefore

\begin{equation}
W_{\mathrm{CD}}
=
m
\odot
\frac{
V_{\mathrm{CD}}
}{
\bar{n}
}.
\label{eq:dora_simple_weight}
\end{equation}

The forward numerical value is unchanged by the stop-gradient operation;
only gradient propagation through the normalization denominator is removed.
We use this detached-denominator configuration for the reported ChainDoRA
experiments.


\subsection{Balanced Nonzero TT Initialization}
\label{subsec:initialization}

Initialization requires additional care for a multiplicative TT
parameterization. Conventional LoRA commonly initializes one low-rank factor
to zero so that the initial residual vanishes. Directly transferring this
strategy to a chain of multiple tensor factors is undesirable because a zero
factor can suppress gradients to other multiplicative factors. A similar
zero-gradient issue for zero-initialized tensor factors has been reported for
TT-based PEFT in LoRETTA~\cite{yang2024loretta}.

ChainDoRA therefore initializes every TT core with nonzero, fan-in-scaled
Gaussian entries. The magnitude vector is initialized from the frozen
pretrained weight according to

\begin{equation}
m^{(0)}
=
\left\|
W_0
\right\|_{\mathrm{row}}.
\label{eq:chaindora_m_init}
\end{equation}

Because products of several tensor cores can develop strongly imbalanced
scales, we next apply gauge-preserving balancing between neighboring TT
cores. For any positive scalar $\lambda_j$, the transformation

\begin{equation}
\mathcal{G}_j
\leftarrow
\lambda_j\mathcal{G}_j,
\qquad
\mathcal{G}_{j+1}
\leftarrow
\lambda_j^{-1}\mathcal{G}_{j+1}
\label{eq:gauge_balance}
\end{equation}

leaves their joint contraction unchanged. We use this freedom to reduce
large scale disparities between adjacent cores.

After contraction, the two directional factors are additionally balanced
across the adapter boundary. Let

\begin{equation}
a_{\mathrm{rms}}
=
\operatorname{RMS}
\left(
A_{\mathrm{TT}}
\right),
\qquad
b_{\mathrm{rms}}
=
\operatorname{RMS}
\left(
B_{\mathrm{TT}}
\right).
\label{eq:ab_rms}
\end{equation}

Choosing

\begin{equation}
\gamma
=
\sqrt{
\frac{
b_{\mathrm{rms}}
}{
a_{\mathrm{rms}}
}
}
\label{eq:boundary_gamma}
\end{equation}

and applying

\begin{equation}
A_{\mathrm{TT}}
\leftarrow
\gamma A_{\mathrm{TT}},
\qquad
B_{\mathrm{TT}}
\leftarrow
\gamma^{-1}B_{\mathrm{TT}}
\label{eq:boundary_balance}
\end{equation}

preserves the directional product,

\begin{equation}
\left(
\gamma^{-1}B_{\mathrm{TT}}
\right)
\left(
\gamma A_{\mathrm{TT}}
\right)
=
B_{\mathrm{TT}}A_{\mathrm{TT}},
\label{eq:boundary_product_invariance}
\end{equation}

while reducing scale imbalance between the two sides of the TT chain.

Unlike exact zero initialization, ChainDoRA starts with a small but nonzero
directional perturbation. We explicitly calibrate its relative magnitude as

\begin{equation}
\delta_0
=
\frac{
\left\|
sB_{\mathrm{TT}}^{(0)}
A_{\mathrm{TT}}^{(0)}
\right\|_{\mathrm{F}}
}{
\left\|
W_0
\right\|_{\mathrm{F}}
}.
\label{eq:initial_relative_update}
\end{equation}

For all experiments, the initialization is rescaled to satisfy

\begin{equation}
\delta_0
=
\tau,
\qquad
\tau=10^{-3}.
\label{eq:initial_update_target}
\end{equation}

Thus, ChainDoRA begins in a controlled neighborhood of the pretrained
transformation while keeping all TT factors nonzero and trainable from the
first optimization step.

The calibration does not require materializing the dense product
$B_{\mathrm{TT}}A_{\mathrm{TT}}$. Its Frobenius norm follows from the
$r\times r$ Gram matrices,

\begin{equation}
\left\|
B_{\mathrm{TT}}A_{\mathrm{TT}}
\right\|_{\mathrm{F}}^2
=
\operatorname{tr}
\left[
\left(
B_{\mathrm{TT}}^{\mathrm{T}}
B_{\mathrm{TT}}
\right)
\left(
A_{\mathrm{TT}}
A_{\mathrm{TT}}^{\mathrm{T}}
\right)
\right].
\label{eq:ba_frobenius_gram}
\end{equation}

This allows initialization calibration to be performed using low-dimensional
intermediate quantities.

Algorithm~\ref{alg:chaindora} summarizes the complete ChainDoRA training
pipeline, including TT-core initialization, construction of the directional
factors, factored row normalization, parameter optimization, and final weight
merging.

\begin{algorithm}[tb]
\caption{ChainDoRA Training and Weight Merging}
\label{alg:chaindora}
\begin{algorithmic}[1]
\STATE \textbf{Input:} Frozen weights $\{W_0^{(\ell)}\}$, data $\mathcal{D}$,
adapter rank $r$, TT rank $\rho$, scale $\alpha$, target $\tau$
\STATE $s \leftarrow \alpha/r$
\STATE \textbf{Initialization}
\FOR{each adapted layer $\ell$}
    \STATE Tensorize $d_{\mathrm{in}}^{(\ell)}$ and
    $d_{\mathrm{out}}^{(\ell)}$
    \STATE Set $\mathbf{R}=(1,\rho,\ldots,\rho,r,\rho,\ldots,\rho,1)$
    \STATE Initialize all $\{\mathcal{G}_{j}^{(\ell)}\}_{j=1}^{N}$
    with nonzero fan-in-scaled Gaussian entries
    \STATE $m^{(\ell)}\leftarrow\|W_0^{(\ell)}\|_{\mathrm{row}}$
    \STATE Gauge-balance neighboring TT cores and contract them into
    $A_{\mathrm{TT}}^{(\ell)}$ and $B_{\mathrm{TT}}^{(\ell)}$
    \STATE Balance the two factors across the rank-$r$ boundary
    \STATE Calibrate
    $\delta_0^{(\ell)}=
    \|sB_{\mathrm{TT}}^{(\ell)}A_{\mathrm{TT}}^{(\ell)}\|_F/
    \|W_0^{(\ell)}\|_F \approx \tau$
\ENDFOR
\STATE \textbf{Fine-tuning}
\FOR{each mini-batch $(x,y)\in\mathcal{D}$}
    \FOR{each adapted layer $\ell$}
        \STATE Contract TT cores into
        $A_{\mathrm{TT}}^{(\ell)}$ and $B_{\mathrm{TT}}^{(\ell)}$
        \STATE $\Delta y^{(\ell)}\leftarrow
        sB_{\mathrm{TT}}^{(\ell)}(A_{\mathrm{TT}}^{(\ell)}x)$
        \STATE Compute factored row norm $n^{(\ell)}$ using
        Eqs.~(\ref{eq:q_term})--(\ref{eq:factored_row_norm})
        \STATE $\bar n^{(\ell)}\leftarrow
        \operatorname{stopgrad}(n^{(\ell)})$
        \STATE $y^{(\ell)}\leftarrow
        (m^{(\ell)}\oslash\bar n^{(\ell)})\odot
        [W_0^{(\ell)}x+\Delta y^{(\ell)}]$
    \ENDFOR
    \STATE Compute $\mathcal{L}_{\mathrm{resp}}$ and update only
    $\{m^{(\ell)},\mathcal{G}_{1:N}^{(\ell)}\}$ with AdamW
\ENDFOR
\STATE \textbf{Weight merging}
\FOR{each adapted layer $\ell$}
    \STATE Contract final TT cores into
    $A_{\mathrm{TT}}^{(\ell)}$ and $B_{\mathrm{TT}}^{(\ell)}$
    \STATE $V_{\mathrm{CD}}^{(\ell)}\leftarrow W_0^{(\ell)}
    +sB_{\mathrm{TT}}^{(\ell)}A_{\mathrm{TT}}^{(\ell)}$
    \STATE $W_{\mathrm{CD}}^{(\ell)}\leftarrow
    m^{(\ell)}\odot V_{\mathrm{CD}}^{(\ell)}
    /\|V_{\mathrm{CD}}^{(\ell)}\|_{\mathrm{row}}$
\ENDFOR
\STATE \textbf{Return:} $\{W_{\mathrm{CD}}^{(\ell)}\}$
\end{algorithmic}
\end{algorithm}

\subsection{Trainable-Parameter Complexity}
\label{subsec:parameter_complexity}

For a rank-$r$ LoRA adapter, the number of trainable parameters associated
with one adapted matrix is

\begin{equation}
P_{\mathrm{LoRA}}
=
r
\left(
d_{\mathrm{in}}
+
d_{\mathrm{out}}
\right).
\label{eq:lora_parameter_count}
\end{equation}

DoRA additionally learns a magnitude parameter for each output row, giving

\begin{equation}
P_{\mathrm{DoRA}}
=
r
\left(
d_{\mathrm{in}}
+
d_{\mathrm{out}}
\right)
+
d_{\mathrm{out}}.
\label{eq:dora_parameter_count}
\end{equation}

For ChainDoRA, the dense low-rank matrices are replaced by the TT cores.
The number of trainable TT parameters is the standard TT storage cost

\begin{equation}
P_{\mathrm{TT}}
=
\sum_{j=1}^{N}
R_{j-1}
n_j
R_j.
\label{eq:tt_parameter_count}
\end{equation}

Including the trainable magnitude vector,

\begin{equation}
P_{\mathrm{ChainDoRA}}
=
d_{\mathrm{out}}
+
\sum_{j=1}^{N}
R_{j-1}
n_j
R_j.
\label{eq:chaindora_parameter_count}
\end{equation}

The relative parameter ratio with respect to DoRA is therefore

\begin{equation}
\eta_{\mathrm{CD/DoRA}}
=
\frac{
d_{\mathrm{out}}
+
\sum_{j=1}^{N}
R_{j-1}n_jR_j
}{
r(d_{\mathrm{in}}+d_{\mathrm{out}})
+
d_{\mathrm{out}}
}.
\label{eq:chaindora_parameter_ratio}
\end{equation}

Correspondingly, the percentage reduction is

\begin{equation}
\mathcal{R}_{\mathrm{CD/DoRA}}
=
100
\left(
1-\eta_{\mathrm{CD/DoRA}}
\right).
\label{eq:parameter_reduction}
\end{equation}

Equation~(\ref{eq:chaindora_parameter_count}) also makes the role of the
internal TT rank explicit: reducing $\rho$ decreases the TT-core parameter
count without changing the adapter boundary rank $r$. This provides a
direct mechanism for controlling the parameter--capacity trade-off, which is
examined empirically in the TT-rank ablation study.


\subsection{Adapter Merging and Inference}
\label{subsec:adapter_merge}

After fine-tuning, the TT cores are contracted to obtain
$A_{\mathrm{TT}}$ and $B_{\mathrm{TT}}$, and the final effective weight is
computed using (\ref{eq:chaindora_complete}). The resulting matrix
$W_{\mathrm{CD}}$ can then replace the corresponding frozen pretrained
weight. Consequently, the TT adapter does not require a separate inference
branch after merging, analogous to the mergeable property of LoRA and
DoRA~\cite{hu2022lora,liu2024dora}.

The proposed factorization is primarily intended to reduce the number of
trainable and stored task-specific parameters. This reduction should not be
interpreted as an equivalent reduction in training FLOPs, because contracting
TT cores and evaluating the normalized directional update introduce
additional operations. We therefore report parameter efficiency and
predictive performance separately in the experimental analysis.

\section{Experimental Setup}
\label{sec:experiments}

We evaluate ChainDoRA under a controlled parameter-efficient adaptation
setting designed to isolate the effect of the proposed TT-factorized
directional parameterization. All primary experiments use the same
pretrained backbone, adaptation rank, target modules, optimization schedule,
training data, and evaluation pipeline for LoRA, DoRA, and ChainDoRA.
Unless otherwise stated, only the adapter parameterization differs between
methods.


\subsection{Backbone and Training Data}
\label{subsec:backbone_data}

We use LLaMA-7B~\cite{touvron2023llama} as the frozen pretrained backbone.
LLaMA is a decoder-only Transformer family containing models ranging from
7B to 65B parameters~\cite{touvron2023llama}. The 7B configuration is used
throughout the primary experiments to maintain a fixed backbone while
comparing the parameter efficiency of different adaptation methods.

Our implementation follows the commonsense-reasoning adaptation pipeline
used in the public DoRA framework~\cite{liu2024dora}. Rather than training
on the complete commonsense instruction collection, we construct a fixed
adaptation pool containing 15,119 instruction--response examples. The adaptation pool is derived from the Commonsense170K collection introduced with LLM-Adapters~\cite{hu2023llmadapters}.
A validation subset of 120 examples is held out using a fixed random seed of
42, leaving 14,999 examples for parameter optimization. All LoRA, DoRA,
ChainDoRA, and ChainDoRA ablation experiments use exactly the same split.

Each training instance consists of an instruction and its target response.
We employ \emph{response-only supervision}: tokens belonging to the
instruction portion of the prompt are masked from the language-modeling loss,
and only response tokens contribute to optimization. Let
$\mathbf{y}=(y_1,\ldots,y_T)$ denote the tokenized instruction--response
sequence, and let $\mathcal{S}$ denote the token positions corresponding to
the target response. The training objective is

\begin{equation}
\mathcal{L}_{\mathrm{resp}}
=
-
\frac{1}{|\mathcal{S}|}
\sum_{t\in\mathcal{S}}
\log
p_{\theta}
\left(
y_t
\mid
y_{<t}
\right),
\label{eq:response_only_loss}
\end{equation}

where tokens outside $\mathcal{S}$ are ignored when computing the loss.
The same serialized instruction template and response delimiter are used
during fine-tuning and evaluation to avoid train--test differences caused by
prompt formatting.


\subsection{Adaptation Configuration}
\label{subsec:adaptation_configuration}

For the main comparison, LoRA, DoRA, and ChainDoRA use the same adapter rank

\begin{equation}
r=32
\label{eq:exp_adapter_rank}
\end{equation}

and scaling coefficient

\begin{equation}
\alpha=64.
\label{eq:exp_alpha}
\end{equation}

Adaptation is applied to five linear transformations in every Transformer
block,

\begin{equation}
\mathcal{M}_{\mathrm{full}}
=
\{
q_{\mathrm{proj}},
k_{\mathrm{proj}},
v_{\mathrm{proj}},
up_{\mathrm{proj}},
down_{\mathrm{proj}}
\}.
\label{eq:full_target_modules}
\end{equation}

For the 32 Transformer layers of LLaMA-7B, this corresponds to 160 adapted
linear modules.

The main ChainDoRA configuration uses internal TT rank

\begin{equation}
\rho=16.
\label{eq:main_tt_rank}
\end{equation}

The hidden dimension of LLaMA-7B is tensorized as

\begin{equation}
4096
=
16\times16\times16,
\label{eq:hidden_tensorization}
\end{equation}

whereas the feed-forward dimension is represented as

\begin{equation}
11008
=
16\times16\times43.
\label{eq:ffn_tensorization}
\end{equation}

Table~\ref{tab:tensorization} summarizes the resulting mode assignments.


\begin{table}[t]
\centering
\caption{Tensorization used for ChainDoRA on LLaMA-7B.}
\label{tab:tensorization}
\resizebox{\columnwidth}{!}{
\begin{tabular}{lcccc}
\toprule
\textbf{Module}
&
$d_{\mathrm{in}}$
&
$d_{\mathrm{out}}$
&
\textbf{Input Modes}
&
\textbf{Output Modes}
\\
\midrule

$q_{\mathrm{proj}}$
& 4096 & 4096
& $(16,16,16)$
& $(16,16,16)$
\\

$k_{\mathrm{proj}}$
& 4096 & 4096
& $(16,16,16)$
& $(16,16,16)$
\\

$v_{\mathrm{proj}}$
& 4096 & 4096
& $(16,16,16)$
& $(16,16,16)$
\\

$up_{\mathrm{proj}}$
& 4096 & 11008
& $(16,16,16)$
& $(16,16,43)$
\\

$down_{\mathrm{proj}}$
& 11008 & 4096
& $(16,16,43)$
& $(16,16,16)$
\\

\bottomrule
\end{tabular}}
\end{table}

Under this configuration, conventional LoRA contains 56.10M trainable
parameters and DoRA contains 56.98M trainable parameters. ChainDoRA with
$\rho=16$ requires 5.35M trainable parameters while preserving the same
adapter boundary rank $r=32$. Exact parameter-efficiency comparisons are
reported in Section~\ref{sec:results}.


\subsection{Optimization and Implementation Details}
\label{subsec:optimization}

All methods are fine-tuned for three epochs using AdamW. The learning rate is
set to

\begin{equation}
\eta
=
2\times10^{-4},
\label{eq:learning_rate}
\end{equation}

with zero weight decay and 100 warm-up steps. The global batch size and
micro-batch size are both 4, resulting in one optimizer update per
micro-batch on the single-GPU setup. The maximum tokenized sequence length is
256 and the adapter dropout probability is 0.05.

Training uses FP16 precision and gradient checkpointing. Validation is
performed every 80 optimization steps, and adapter checkpoints are saved at
the same interval. At the end of training, the best validation checkpoint is
restored. All experiments are implemented in PyTorch using the
PEFT-compatible DoRA codebase extended with the proposed ChainDoRA module.
Training is conducted on a single NVIDIA GeForce RTX 5090 GPU with 32~GB of
memory.

Table~\ref{tab:training_configuration} summarizes the common training
configuration.


\begin{table}[t]
\centering
\caption{Training configuration used for the primary comparison.}
\label{tab:training_configuration}
\resizebox{\columnwidth}{!}{
\begin{tabular}{lc}
\toprule
\textbf{Configuration} & \textbf{Value} \\
\midrule
Backbone & LLaMA-7B \\
Adaptation pool & 15,119 examples \\
Optimization examples & 14,999 \\
Validation examples & 120 \\
Training epochs & 3 \\
Optimizer & AdamW \\
Learning rate & $2\times10^{-4}$ \\
Weight decay & 0 \\
Warm-up steps & 100 \\
Global batch size & 4 \\
Micro-batch size & 4 \\
Maximum sequence length & 256 \\
Adapter dropout & 0.05 \\
Adapter rank $r$ & 32 \\
Scaling coefficient $\alpha$ & 64 \\
Main ChainDoRA TT rank $\rho$ & 16 \\
Target modules & Q, K, V, Up, Down \\
Validation interval & 80 steps \\
Checkpoint interval & 80 steps \\
Precision & FP16 \\
Gradient checkpointing & Enabled \\
Supervision & Response only \\
Hardware & NVIDIA RTX 5090 (32 GB) \\
\bottomrule
\end{tabular}}
\end{table}


\subsection{Commonsense Reasoning Benchmarks}
\label{subsec:benchmarks}

We evaluate the adapted models on seven established commonsense reasoning
benchmarks covering complementary forms of linguistic, physical, social,
coreference, and science-oriented reasoning.

BoolQ~\cite{clark2019boolq} consists of naturally occurring yes/no questions
and evaluates binary question answering. PIQA~\cite{bisk2020piqa} measures
physical commonsense reasoning through two candidate solutions to everyday
physical situations. SocialIQA~\cite{sap2019socialiqa} evaluates reasoning
about social interactions, intentions, and consequences. WinoGrande
~\cite{sakaguchi2020winogrande} is a large-scale adversarially constructed
Winograd-style benchmark designed to reduce exploitable annotation biases.

ARC~\cite{clark2018arc} consists of natural grade-school science questions
and is divided into ARC-Easy and ARC-Challenge. The Challenge partition was
constructed from questions that were incorrectly answered by baseline
retrieval and co-occurrence systems, making it a more difficult subset.
OpenBookQA~\cite{mihaylov2018openbookqa} evaluates elementary science
question answering that may require combining a provided scientific fact
with broader common knowledge.

Table~\ref{tab:benchmark_summary} lists the seven evaluation sets and the
number of examples used in our experiments.


\begin{table}[t]
\centering
\caption{Commonsense reasoning benchmarks used for evaluation.}
\label{tab:benchmark_summary}
\resizebox{\columnwidth}{!}{
\begin{tabular}{lcl}
\toprule
\textbf{Benchmark}
&
\textbf{Test Examples}
&
\textbf{Primary Reasoning Type}
\\
\midrule

BoolQ
& 3,270
& Yes/no reasoning
\\

PIQA
& 1,838
& Physical commonsense
\\

SocialIQA
& 1,954
& Social commonsense
\\

WinoGrande
& 1,267
& Coreference / commonsense
\\

ARC-Easy
& 2,376
& Science reasoning
\\

ARC-Challenge
& 1,172
& Challenging science reasoning
\\

OpenBookQA
& 500
& Science + common knowledge
\\

\midrule
\textbf{Total}
& \textbf{12,377}
& --
\\

\bottomrule
\end{tabular}}
\end{table}


\subsection{Evaluation Protocol and Metrics}
\label{subsec:evaluation_protocol}

A single evaluation pipeline is used for all methods. Each test example is
converted to an instruction containing the question, candidate answers, and
the required benchmark-specific output identifier. The same instruction
serialization used during fine-tuning is retained during inference, with the
response field left empty for generation.

Before evaluation, the learned adapter parameters are merged into their
corresponding linear transformations. Generation is deterministic and uses
beam search with four beams,

\begin{equation}
N_{\mathrm{beam}}=4,
\label{eq:num_beams}
\end{equation}

with a maximum of 32 newly generated tokens,

\begin{equation}
T_{\mathrm{gen}}^{\max}=32.
\label{eq:max_new_tokens}
\end{equation}

Sampling is disabled. The generated continuation is mapped to the
benchmark-specific answer vocabulary. Examples include
\texttt{true}/\texttt{false} for BoolQ,
\texttt{solution1}/\texttt{solution2} for PIQA,
\texttt{option1}/\texttt{option2} for WinoGrande, and
\texttt{answer1}, \texttt{answer2}, $\ldots$ for the multiple-choice
question-answering datasets. The first valid answer identifier appearing in
the generated response is used as the prediction. A generation containing no
valid identifier is counted as incorrect.

The primary metric for benchmark $j$ is classification accuracy,

\begin{equation}
\mathrm{Acc}_j
=
\frac{
N_{\mathrm{correct},j}
}{
N_{\mathrm{total},j}
}
\times100.
\label{eq:benchmark_accuracy}
\end{equation}

To summarize performance across the complete benchmark suite, we report the
unweighted macro-average over the seven tasks,

\begin{equation}
\mathrm{Avg.}
=
\frac{1}{7}
\sum_{j=1}^{7}
\mathrm{Acc}_j.
\label{eq:seven_task_average}
\end{equation}

The macro-average gives equal weight to each benchmark regardless of the
number of test examples.


\subsection{Ablation Study Design}
\label{subsec:ablation_setup}

We conduct two principal ablation studies to examine the effects of TT
capacity and adapter placement.

First, the internal TT rank is varied while keeping the adapter rank fixed at
$r=32$:

\begin{equation}
\rho
\in
\{4,8,16\}.
\label{eq:tt_rank_ablation}
\end{equation}

All five projection types in
$\mathcal{M}_{\mathrm{full}}$ remain adapted in this experiment. This
ablation isolates the effect of TT representation capacity without changing
the nominal rank of the directional update.

Second, we vary the locations at which ChainDoRA is inserted. The evaluated
target sets are

\begin{equation}
\mathcal{M}_{\mathrm{QV}}
=
\{
q_{\mathrm{proj}},
v_{\mathrm{proj}}
\},
\label{eq:qv_modules}
\end{equation}

\begin{equation}
\mathcal{M}_{\mathrm{QKV}}
=
\{
q_{\mathrm{proj}},
k_{\mathrm{proj}},
v_{\mathrm{proj}}
\},
\label{eq:qkv_modules}
\end{equation}

\begin{equation}
\mathcal{M}_{\mathrm{MLP}}
=
\{
up_{\mathrm{proj}},
down_{\mathrm{proj}}
\},
\label{eq:mlp_modules}
\end{equation}

and

\begin{equation}
\mathcal{M}_{\mathrm{QV+MLP}}
=
\{
q_{\mathrm{proj}},
v_{\mathrm{proj}},
up_{\mathrm{proj}},
down_{\mathrm{proj}}
\}.
\label{eq:qv_mlp_modules}
\end{equation}

These configurations are compared with full QKV+MLP adaptation. Unless
otherwise specified, the projection-location study uses $\rho=16$.

We additionally evaluate a highly compressed configuration combining
$\rho=8$ with MLP-only adaptation. This variant is included to examine
whether reductions in TT capacity and adaptation coverage can be combined
while retaining useful downstream performance.


\begin{table}[t]
\centering
\caption{ChainDoRA configurations used in the ablation studies.}
\label{tab:ablation_configurations}
\resizebox{\columnwidth}{!}{
\begin{tabular}{lccr}
\toprule
\textbf{Configuration}
&
$\rho$
&
\textbf{Target Set}
&
\textbf{Trainable Params.}
\\
\midrule

TT16-Full
& 16 & QKV+MLP & 5,346,816
\\

TT8-Full
& 8 & QKV+MLP & 2,784,000
\\

TT4-Full
& 4 & QKV+MLP & 1,748,352
\\

TT16-QV
& 16 & QV & 1,867,776
\\

TT16-QKV
& 16 & QKV & 2,801,664
\\

TT16-MLP
& 16 & MLP & 2,545,152
\\

TT16-QV+MLP
& 16 & QV+MLP & 4,412,928
\\

TT8-MLP
& 8 & MLP & 1,383,168
\\

\bottomrule
\end{tabular}}
\end{table}


\subsection{Cross-Architecture Parameter-Scaling Analysis}
\label{subsec:scaling_setup}

To evaluate how the parameterization scales beyond LLaMA-7B, we additionally
perform a configuration-level parameter analysis across model families and
sizes. This analysis does not require loading pretrained model weights or
performing additional fine-tuning. Instead, architectural configurations are
used to determine the dimensions and number of target linear transformations,
after which the trainable parameter counts of LoRA, DoRA, and ChainDoRA are
computed analytically.

The analysis includes Qwen2.5-3B and Qwen2.5-72B
~\cite{qwen2024qwen25}, Mistral-7B~\cite{jiang2023mistral},
Phi-3-Medium~\cite{abdin2024phi3}, OPT-30B~\cite{zhang2022opt},
and Falcon-40B~\cite{almazrouei2023falcon}. These architectures span
approximately 3B to 72B parameters and represent multiple Transformer
families with different attention and feed-forward configurations. 
For each architecture, target modules are selected according to their native attention and
feed-forward projection structure. Full-target ChainDoRA is used for the
primary apples-to-apples comparison with LoRA and DoRA, while selective
MLP-only variants are reported separately as additional compression
configurations.

This experiment is intended to characterize parameter scaling rather than
downstream accuracy across these model families; no pretrained weights are
loaded and no performance claims are made for the configuration-only models.

\section{Results and Discussion}
\label{sec:results}

This section evaluates ChainDoRA from three complementary perspectives:
predictive performance, trainable-parameter efficiency, and sensitivity to
the design of the tensorized adapter. We first compare ChainDoRA with matched
LoRA and DoRA baselines on the seven commonsense reasoning benchmarks.
We then study the effect of the internal TT rank and the placement of
ChainDoRA across attention and feed-forward projections. Finally, we examine
the parameter scaling of the proposed representation across several
large-language-model architectures.


\subsection{Overall Commonsense Reasoning Performance}
\label{subsec:main_results}

Table~\ref{tab:main_results} reports the primary comparison among LoRA,
DoRA, and ChainDoRA. All three methods use the same LLaMA-7B backbone,
rank $r=32$, scaling coefficient $\alpha=64$, target modules, optimization
schedule, response-only training objective, and evaluation protocol.
ChainDoRA uses the full QKV+MLP target set with internal TT rank
$\rho=16$.


\begin{table*}[t]
\centering
\caption{Commonsense reasoning performance of LoRA, DoRA, and ChainDoRA
under the controlled 15,119-example adaptation setting. Accuracy is reported
in percent. The best accuracy in each benchmark column is shown in bold.}
\label{tab:main_results}
\resizebox{\textwidth}{!}{
\begin{tabular}{lrrrrrrrrr}
\toprule
\textbf{Method}
&
\textbf{Params.}
&
\textbf{BoolQ}
&
\textbf{PIQA}
&
\textbf{SocialIQA}
&
\textbf{WinoGrande}
&
\textbf{ARC-E}
&
\textbf{ARC-C}
&
\textbf{OBQA}
&
\textbf{Avg.}
\\
\midrule

LoRA
& 56.10M
& 64.83
& 72.09
& \textbf{72.47}
& 65.82
& 78.03
& 62.88
& 73.00
& 69.88
\\

DoRA
& 56.98M
& 65.60
& 74.54
& 70.21
& 65.51
& 76.60
& 60.07
& 73.20
& 69.39
\\

\textbf{ChainDoRA}
& \textbf{5.35M}
& \textbf{66.24}
& \textbf{77.64}
& 71.80
& \textbf{66.30}
& \textbf{82.32}
& \textbf{67.58}
& \textbf{74.20}
& \textbf{72.30}
\\

\bottomrule
\end{tabular}}
\end{table*}

ChainDoRA achieves the highest macro-average accuracy, reaching
72.30\% across the seven benchmarks. This corresponds to improvements
of 2.42 percentage points over LoRA and 2.91 percentage points over
DoRA. Relative to LoRA, ChainDoRA achieves higher accuracy on six of
the seven benchmarks, with particularly notable improvements on PIQA,
ARC-Easy, and ARC-Challenge. The respective gains are 5.55, 4.29,
and 4.70 percentage points. LoRA obtains the highest SocialIQA score,
exceeding ChainDoRA by 0.67 percentage points.

Compared with DoRA, ChainDoRA achieves higher accuracy on all seven
tasks. The largest difference is observed on ARC-Challenge, where
accuracy increases from 60.07\% to 67.58\%, corresponding to a
7.51-percentage-point improvement. ARC-Easy increases from 76.60\%
to 82.32\%, a gain of 5.72 percentage points. These results show that,
under the controlled adaptation setting considered here, the connected
TT parameterization achieves higher predictive performance while
substantially reducing the number of trainable parameters.

DoRA obtains a slightly lower average accuracy than LoRA in our
experiments (69.39\% versus 69.88\%). This observation is specific to
the matched training and evaluation configuration used in this work.
The original DoRA study reports improvements over LoRA across multiple
language and multimodal settings~\cite{liu2024dora}; therefore, the
present result should not be interpreted as a general comparison of
the learning capabilities of LoRA and DoRA. Instead, the matched
baselines establish the relative performance of ChainDoRA within the
controlled response-only adaptation regime examined in this study.


\subsection{Parameter Efficiency}
\label{subsec:parameter_efficiency_results}

The predictive improvement of ChainDoRA is accompanied by a substantial
reduction in trainable parameters. Table~\ref{tab:param_efficiency_main}
summarizes the parameter requirements of the three primary methods.

\begin{table}[t]
\centering
\caption{Trainable-parameter efficiency of the primary methods. The
compression factor indicates how many times larger the corresponding
baseline adapter is than ChainDoRA.}
\label{tab:param_efficiency_main}
\resizebox{\columnwidth}{!}{
\begin{tabular}{lrrr}
\toprule
\textbf{Method}
&
\textbf{Trainable Params.}
&
\textbf{Compression}
&
\textbf{Reduction}
\\
\midrule

LoRA
&
56,098,816
&
$10.49\times$
&
90.47\%
\\

DoRA
&
56,975,360
&
$10.66\times$
&
90.62\%
\\

ChainDoRA
&
5,346,816
&
$1.00\times$
&
--
\\

\bottomrule
\end{tabular}}
\end{table}

At identical adapter rank $r=32$, ChainDoRA reduces the trainable parameter
count by 90.47\% relative to LoRA and 90.62\% relative to DoRA. Equivalently,
LoRA and DoRA require approximately $10.49\times$ and $10.66\times$ more
trainable parameters, respectively. Importantly, this reduction is achieved
without decreasing the nominal rank bound of the directional update: all
three methods use $r=32$, while ChainDoRA controls the complexity of the
factor construction independently through the TT rank $\rho$.

\begin{figure}[t]
    \centering
    \includegraphics[width=\columnwidth]
    {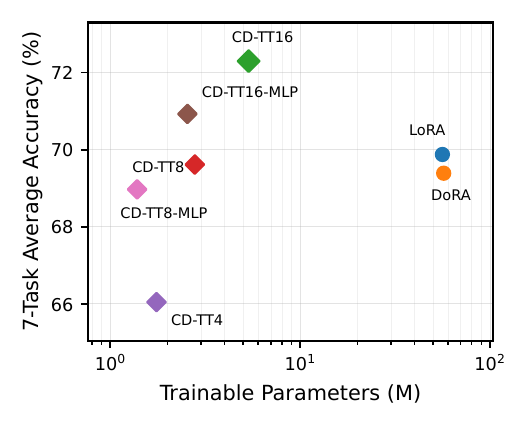}
    \caption{Parameter--accuracy trade-off among LoRA, DoRA, and
    representative ChainDoRA configurations.}
    \label{fig:accuracy_parameter_tradeoff}
\end{figure}

The relationship between trainable parameter count and average benchmark
accuracy is visualized in Fig.~\ref{fig:accuracy_parameter_tradeoff}.
ChainDoRA-TT16 occupies a favorable region of this parameter--accuracy space,
combining the highest average accuracy among the primary methods with a
substantially smaller task-specific parameter footprint.


\subsection{Effect of Tensor-Train Rank}
\label{subsec:tt_rank_results}

The internal TT rank $\rho$ determines the capacity of the connected tensor
representation without altering the adapter boundary rank $r$. To evaluate
this effect, we vary

\begin{equation}
\rho\in\{4,8,16\},
\label{eq:result_tt_ranks}
\end{equation}

while keeping $r=32$ and adapting all QKV+MLP projections.

\begin{table*}[t]
\centering
\caption{Effect of the internal TT rank on ChainDoRA under full QKV+MLP
adaptation with a fixed adapter rank of $r=32$. Accuracy is reported in
percent. The best accuracy in each benchmark column is shown in bold.}
\label{tab:tt_rank_results}
\resizebox{\textwidth}{!}{
\begin{tabular}{crrrrrrrrr}
\toprule
$\boldsymbol{\rho}$
&
\textbf{Params.}
&
\textbf{BoolQ}
&
\textbf{PIQA}
&
\textbf{SocialIQA}
&
\textbf{WinoGrande}
&
\textbf{ARC-E}
&
\textbf{ARC-C}
&
\textbf{OBQA}
&
\textbf{Avg.}
\\
\midrule

4
& 1.748M
& 57.16
& 75.63
& 66.33
& 59.91
& 77.90
& 58.62
& 66.80
& 66.05
\\

8
& 2.784M
& 64.62
& 76.61
& 67.96
& 63.85
& 80.98
& 62.29
& 71.00
& 69.61
\\

16
& 5.347M
& \textbf{66.24}
& \textbf{77.64}
& \textbf{71.80}
& \textbf{66.30}
& \textbf{82.32}
& \textbf{67.58}
& \textbf{74.20}
& \textbf{72.30}
\\

\bottomrule
\end{tabular}}
\end{table*}

Table~\ref{tab:tt_rank_results} shows a monotonic
capacity--performance relationship over the evaluated ranks. Increasing
$\rho$ from 4 to 8 improves the seven-task average from 66.05\% to
69.61\%, and increasing $\rho$ to 16 further raises it to 72.30\%.

Reducing $\rho$ from 16 to 8 lowers the trainable parameter count from
5.347M to 2.784M, corresponding to a 47.93\% reduction, while reducing
average accuracy by 2.69 percentage points. At $\rho=4$, the adapter
uses 1.748M trainable parameters, a 67.30\% reduction relative to
TT16-Full, with a 6.25-percentage-point decrease in average accuracy.

These results empirically support the intended separation between the
adapter rank $r$ and the TT capacity parameter $\rho$. The former fixes the
maximum rank of the resulting directional update, whereas the latter provides
an independent mechanism for adjusting the expressiveness and storage cost
of its tensorized construction. Similar rank--compression trade-offs have
been observed in previous TT-based adaptation studies
~\cite{yang2024loretta,anjum2024ttlora}; however, in ChainDoRA this trade-off
occurs specifically within the directional component of magnitude--direction
adaptation.

\begin{figure}[t]
    \centering
    \includegraphics[width=\columnwidth]
    {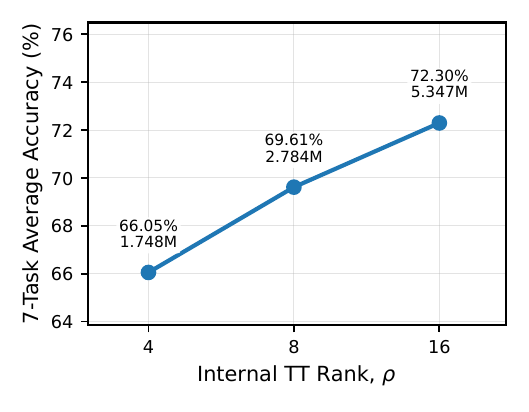}
    \caption{Effect of internal TT rank $\rho$ on the seven-task average
    accuracy and ChainDoRA parameter count. Increasing $\rho$ improves predictive
    performance while increasing the size of the TT representation.}
    \label{fig:tt_rank_ablation}
\end{figure}

\subsection{Effect of Adaptation Location}
\label{subsec:projection_ablation_results}

We next investigate whether the full set of attention and feed-forward
projections is necessary for effective ChainDoRA adaptation. Table
~\ref{tab:projection_ablation} compares full QKV+MLP adaptation with
selective insertion into QV, QKV, MLP, and QV+MLP projections. All variants
in this experiment use $\rho=16$.

\begin{table}[t]
\centering
\caption{Effect of ChainDoRA adapter placement.}
\label{tab:projection_ablation}
\small
\setlength{\tabcolsep}{7pt}
\renewcommand{\arraystretch}{1.08}
\begin{tabular}{@{}lccc@{}}
\toprule
\textbf{Target} &
\textbf{Params. (M)} &
\textbf{Avg. (\%)} &
\boldmath$\Delta$\textbf{ (pp)} \\
\midrule
QV       & 1.868 & 70.43 & $-1.87$ \\
QKV      & 2.802 & 68.60 & $-3.70$ \\
MLP      & 2.545 & 70.93 & $-1.37$ \\
QV+MLP   & 4.413 & 70.93 & $-1.37$ \\
\midrule
QKV+MLP  & 5.347 & \textbf{72.30} & -- \\
\bottomrule
\end{tabular}
\end{table}

Full QKV+MLP adaptation provides the highest overall accuracy. However,
selective adaptation reveals that a substantial portion of the performance
can be retained with considerably fewer parameters. In particular, MLP-only
adaptation achieves 70.93\% average accuracy with 2.545M trainable
parameters, reducing the adapter size by 52.40\% relative to TT16-Full while
losing only 1.37 percentage points.

The QV+MLP configuration also achieves 70.93\% average accuracy,
matching the MLP-only configuration at the reported precision, but
requires 4.413M trainable parameters. Thus, adding Q and V projections
increases the parameter count by approximately 1.868M without improving
the reported macro-average. Among the selective configurations, MLP-only
adaptation therefore provides the strongest parameter--accuracy balance.

Interestingly, QKV adaptation obtains a lower average accuracy than QV
adaptation despite using more trainable parameters. This result indicates
that increasing the number of adapted projection types does not necessarily
produce monotonic performance improvements under a fixed TT configuration.
The effectiveness of the adapter therefore depends not only on its total
parameter count but also on where the structured update is introduced.

\begin{figure*}[t]
    \centering
    \includegraphics[width=0.90\textwidth]
    {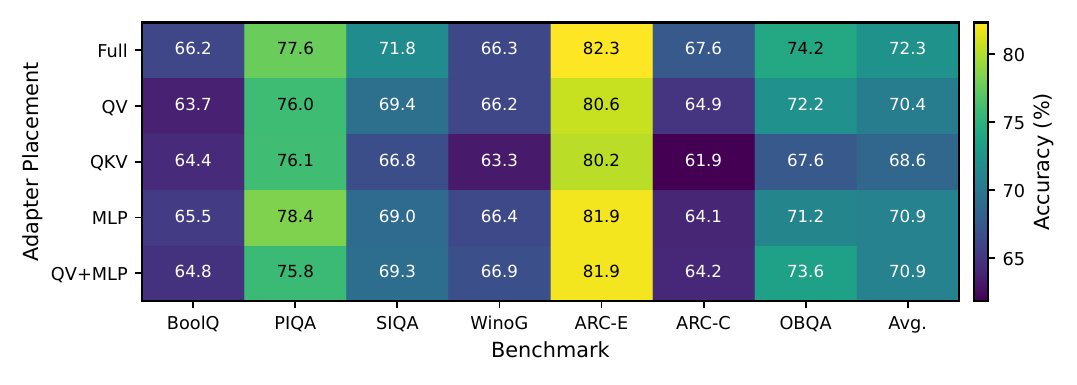}
    \caption{Per-benchmark performance of ChainDoRA under different
    adapter-placement strategies.}
    \label{fig:projection_heatmap}
\end{figure*}


\subsection{Combined Rank and Placement Compression}
\label{subsec:combined_compression}

The TT-rank and projection-location experiments suggest two independent ways
to reduce ChainDoRA's parameter count: decreasing the internal TT rank and
restricting adaptation to a subset of projections. We therefore evaluate a
combined configuration using $\rho=8$ and MLP-only adaptation.

Table~\ref{tab:projection_ablation} compares full QKV+MLP adaptation with
selective placement of ChainDoRA. Here, $\Delta$ denotes the change in
seven-task average accuracy, in percentage points, relative to the full
QKV+MLP configuration.

\begin{table}[t]
\centering
\caption{Representative ChainDoRA parameter--accuracy trade-offs.}
\label{tab:efficiency_frontier}
\small
\setlength{\tabcolsep}{9pt}
\renewcommand{\arraystretch}{1.08}
\begin{tabular}{@{}lcc@{}}
\toprule
\textbf{Configuration} &
\textbf{Params. (M)} &
\textbf{Avg. (\%)} \\
\midrule
TT16-Full & 5.347 & 72.30 \\
TT16-MLP  & 2.545 & 70.93 \\
TT8-Full  & 2.784 & 69.61 \\
TT8-MLP   & 1.383 & 68.97 \\
TT4-Full  & 1.748 & 66.05 \\
\bottomrule
\end{tabular}
\end{table}

The TT8-MLP variant requires only 1.383M trainable parameters,
representing a 74.13\% reduction relative to TT16-Full. Compared with
TT8-Full, MLP-only placement removes 50.32\% of the trainable parameters
while reducing average accuracy by only 0.64 percentage points.

Compared with TT4-Full, TT8-MLP uses 20.89\% fewer parameters while
achieving a 2.92-percentage-point higher average accuracy. Thus,
TT8-MLP strictly dominates TT4-Full with respect to both parameter
count and average accuracy within the evaluated configuration space.


\subsection{Cross-Architecture Parameter Scaling}
\label{subsec:parameter_scaling_results}

The preceding experiments evaluate downstream accuracy on LLaMA-7B. To
determine whether the parameter savings are specific to this architecture,
we additionally perform the configuration-level analysis described in
Section~\ref{subsec:scaling_setup}. This experiment calculates trainable
parameter counts from model configurations without loading pretrained weights
and therefore measures structural scaling only.

Table~\ref{tab:cross_arch_scaling} compares full-target DoRA with full-target
ChainDoRA-TT16 across six representative architectures.
Here, CD-TT16 denotes the full-target ChainDoRA configuration with
internal TT rank $\rho=16$.
\begin{table}[t]
\centering
\caption{Cross-architecture trainable-parameter scaling.}
\label{tab:cross_arch_scaling}
\footnotesize
\setlength{\tabcolsep}{3.6pt}
\renewcommand{\arraystretch}{1.08}

\begin{tabular}{@{}lrrrr@{}}
\toprule

\textbf{Model}
&
\multicolumn{2}{c}{\textbf{Params. (M)}}
&
\textbf{Red.}
&
\textbf{Comp.}
\\

\cmidrule(lr){2-3}

&
\textbf{DoRA}
&
\textbf{CD-TT16}
&
\textbf{(\%)}
&
\textbf{($\times$)}
\\

\midrule

Qwen2.5-3B
& 40.67
& 4.97
& 87.77
& 8.18
\\

Mistral-7B
& 57.41
& 5.46
& 90.48
& 10.51
\\

Phi-3-Medium
& 100.25
& 6.09
& 93.92
& 16.45
\\

OPT-30B
& 178.91
& 12.42
& 93.06
& 14.40
\\

Falcon-40B
& 193.72
& 12.12
& 93.74
& 15.98
\\

Qwen2.5-72B
& 286.30
& 19.75
& 93.10
& 14.49
\\

\bottomrule
\end{tabular}
\end{table}

ChainDoRA-TT16 requires fewer trainable parameters than DoRA for every
architecture considered. The compression factor ranges from $8.18\times$ on
Qwen2.5-3B to $16.45\times$ on Phi-3-Medium, with an average compression
factor of approximately $13.34\times$ across the six configurations.
The absolute savings become increasingly large for the larger models; for
example, the Qwen2.5-72B configuration requires 286.30M DoRA parameters but
only 19.75M ChainDoRA-TT16 parameters under the corresponding target-module
mapping.
\begin{figure*}[t]
    \centering
    \includegraphics[width=0.90\textwidth]
    {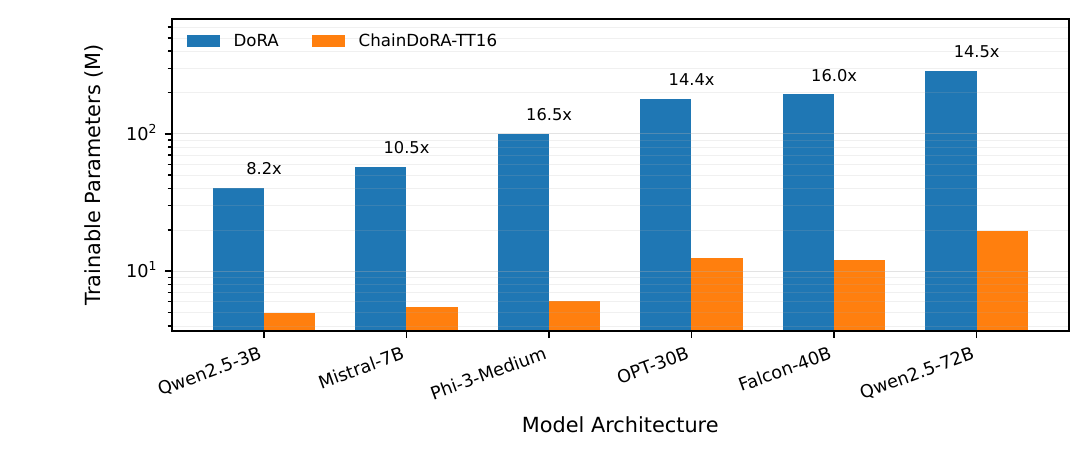}
    \caption{Trainable-parameter scaling of full-target DoRA and
    ChainDoRA-TT16 across representative language-model architectures.}
    \label{fig:cross_arch_scaling}
\end{figure*}
More aggressive selective configurations provide further reductions.
For example, the TT8-MLP variant yields compression factors ranging from
approximately $28.9\times$ to $43.7\times$ relative to full-target DoRA
across the same architecture set. Because these configurations adapt fewer
modules, they are treated separately from the primary full-target comparison.

\subsection{Discussion and Limitations}
\label{subsec:discussion}


The experimental results reveal three main properties of ChainDoRA.
First, the connected TT representation can substantially reduce the
parameter cost of DoRA-style directional adaptation without requiring a
smaller adapter boundary rank. The primary TT16 configuration uses only
5.35M trainable parameters while obtaining the highest average accuracy in
the controlled seven-task comparison. This suggests that a densely
parameterized pair of directional factors is not necessary to obtain an
effective rank-$32$ update in the studied setting.

Second, parameter efficiency is controlled by more than a single scalar
compression parameter. The TT-rank experiment shows that $\rho$ provides a
direct capacity--accuracy trade-off, while the projection study demonstrates
that adapter placement also has a substantial effect. In particular,
MLP-only adaptation retains much of the performance of full adaptation with
approximately half of the trainable parameters. Combining these two design
dimensions produces the TT8-MLP configuration, which improves upon TT4-Full
in both parameter count and accuracy. These observations motivate treating
TT rank and target-module placement as complementary design variables rather
than optimizing either independently.

Third, the configuration-level scaling experiment indicates that the
parameter savings are not restricted to the LLaMA-7B dimensions. The
full-target ChainDoRA construction remains substantially smaller than DoRA
across architectures ranging from approximately 3B to 72B parameters.
However, this experiment evaluates parameter counts only and should not be
interpreted as evidence that the LLaMA-7B accuracy gains transfer directly
to these other model families.

Several limitations remain. The downstream experiments are conducted using a
single pretrained backbone and a fixed 15,119-example adaptation pool rather
than exhaustive full-dataset fine-tuning. The reported conclusions should
therefore be interpreted as evidence from a controlled parameter-efficient
adaptation setting rather than as claims of state-of-the-art commonsense
reasoning performance. In addition, the cross-architecture study is
analytical and does not include model training. The primary baselines use
matched optimization hyperparameters to isolate adapter parameterization;
method-specific hyperparameter searches could change their absolute
performance. Future work should evaluate
ChainDoRA across additional model families, larger adaptation datasets,
multiple random seeds, and broader reasoning and generation benchmarks.
Further investigation of TT contraction kernels and training-time efficiency
would also be valuable, since reductions in trainable parameter count do not
necessarily translate directly into equivalent reductions in computational
cost.
\section{Conclusion}
\label{sec:conclusion}

This work introduced \emph{ChainDoRA}, a parameter-efficient adaptation
framework that combines magnitude--direction weight decomposition with a
connected Tensor-Train representation of the directional low-rank factors.
Instead of directly optimizing the dense matrices used by LoRA and DoRA,
ChainDoRA constructs
$A_{\mathrm{TT}}\in\mathbb{R}^{r\times d_{\mathrm{in}}}$ and
$B_{\mathrm{TT}}\in\mathbb{R}^{d_{\mathrm{out}}\times r}$
from contractions of a shared TT chain. The adapter rank $r$ is retained as
the boundary rank between the input- and output-side contractions, while an
independent internal TT rank $\rho$ controls the capacity and parameter cost
of the structured representation.

Under a controlled 15,119-example adaptation setting with LLaMA-7B,
ChainDoRA-TT16 achieved a seven-task average commonsense reasoning accuracy
of 72.30\%, compared with 69.88\% for LoRA and 69.39\% for DoRA. At the same
adapter rank $r=32$, ChainDoRA required only 5.35M trainable parameters,
compared with 56.10M for LoRA and 56.98M for DoRA, corresponding to a
90.62\% reduction relative to DoRA. The TT-rank ablation further demonstrated
a monotonic capacity--performance trade-off across
$\rho\in\{4,8,16\}$, while the adapter-placement study showed that selective
MLP adaptation can retain much of the full QKV+MLP ChainDoRA performance with
substantially fewer trainable parameters. Combining these two compression
dimensions produced the TT8-MLP configuration, which achieved a more
favorable parameter--accuracy operating point than TT4-Full.

The configuration-level scaling analysis further showed that the structural
parameter savings of ChainDoRA persist across model architectures ranging
from approximately 3B to 72B parameters, although downstream performance on
these additional architectures remains to be evaluated. Overall, the results
suggest that the dense directional factors used in conventional
magnitude--direction adaptation can be replaced by a structured tensor
representation while retaining strong downstream performance and
substantially reducing the number of task-specific trainable parameters.

Future work will investigate ChainDoRA across additional language-model
families, larger and more diverse adaptation datasets, multiple random seeds,
and broader reasoning and generation tasks. Further optimization of TT
contraction and normalization kernels may also improve training-time
efficiency, since reductions in trainable parameter count do not necessarily
translate directly into proportional reductions in computational cost.

% ...

\bibliographystyle{unsrtnat}
\bibliography{Reference}

\end{document}